\documentclass[nohyperref]{article}

\usepackage[accepted]{style/icml2023}
\usepackage[utf8]{inputenc} %
\usepackage[T1]{fontenc}    %
\usepackage{hyperref}       %

\hypersetup{
    colorlinks = true,
    citecolor=blue,
    linkcolor=blue,
    urlcolor=blue
}
\usepackage{url}            %
\usepackage{booktabs}       %
\usepackage{amsfonts}       %
\usepackage{nicefrac}       %
\usepackage{microtype}      %
\usepackage{xcolor}         %
\usepackage{multirow}

\usepackage{graphicx}

\usepackage{parskip}
\usepackage[compact]{titlesec}

\usepackage{pifont}%
\usepackage{amssymb, mathtools, amsmath, amsthm}

\usepackage{tcolorbox}
\usepackage[bf, small]{caption}
\usepackage[inline]{enumitem}
\usepackage{wrapfig}

\usepackage[capitalize,noabbrev]{cleveref}

\usepackage{subfigure}
\usepackage{arydshln}
\definecolor{darkgreen}{rgb}{0.0, 0.5, 0.0}
\definecolor{blue}{rgb}{0.0, 0.47, 0.75}
\definecolor{dartmouthgreen}{rgb}{0.05, 0.5, 0.06}
\definecolor{drab}{rgb}{0.59, 0.44, 0.09}
\definecolor{navyblue}{rgb}{0.0, 0.0, 0.5}

\definecolor{darkgreen}{rgb}{0.0, 0.5, 0.0}
\definecolor{blue}{rgb}{0.0, 0.47, 0.75}
\definecolor{dartmouthgreen}{rgb}{0.05, 0.5, 0.06}
\definecolor{drab}{rgb}{0.59, 0.44, 0.09}
\definecolor{navyblue}{rgb}{0.0, 0.0, 0.5}

\newcommand{\bR}{\mathbb{R}}

\newlist{todolist}{itemize}{2}
\setlist[todolist]{label=$\square$}

\newcommand{\cmark}{{\color{dartmouthgreen} \ding{51}}}
\newcommand{\xmark}{{\color{red} \ding{55}}}

\theoremstyle{plain}

\theoremstyle{definition}

\theoremstyle{remark}

\icmltitlerunning{}

\begin{document}

\twocolumn[
\icmltitle{Unifying Molecular and Textual Representations via Multi-task \\ Language Modelling}



\icmlsetsymbol{equal}{*}

\begin{icmlauthorlist}
\icmlauthor{Dimitrios Christofidellis}{equal,ibm}
\icmlauthor{Giorgio Giannone}{equal,ibm,dtu,mit} \\
\icmlauthor{}{} \\
\icmlauthor{Jannis Born}{ibm,eth}
\icmlauthor{Ole Winther}{dtu,ku}
\icmlauthor{Teodoro Laino}{ibm}
\icmlauthor{Matteo Manica}{ibm}
%\icmlauthor{}{} \\
%\icmlauthor{}{} \\
% \href{https://huggingface.co/spaces/GT4SD/text-chem-t5}{Text-Chem T5 Demo}
\end{icmlauthorlist}

\icmlaffiliation{dtu}{Technical University of Denmark}
\icmlaffiliation{eth}{ETH Zurich}
\icmlaffiliation{ibm}{IBM Research Europe}
\icmlaffiliation{ku}{University of Copenhagen}
\icmlaffiliation{mit}{Massachusetts Institute of Technology}

\icmlcorrespondingauthor{Dimitrios Christofidellis}{dic@zurich.ibm.com}
\icmlcorrespondingauthor{Giorgio Giannone}{gigi@dtu.dk}
\icmlcorrespondingauthor{Matteo Manica}{tte@zurich.ibm.com}

%\icmlkeywords{multitask language models}
\vskip 0.3in 
]

\printAffiliationsAndNotice{\icmlEqualContribution} 

\label{section:abstract}

\begin{abstract}
    The recent advances in neural language models have also been successfully applied to the field of chemistry, offering generative solutions for classical problems in molecular design and synthesis planning. 
    These new methods have the potential to fuel a new era of data-driven automation in scientific discovery. 
    However, specialized models are still typically required for each task, leading to the need for problem-specific fine-tuning and neglecting task interrelations. The main obstacle in this field is the lack of a unified representation between natural language and chemical representations, complicating and limiting human-machine interaction.
    Here, we propose the first multi-domain, multi-task language model that can solve a wide range of tasks in both the chemical and natural language domains. 
    Our model can handle chemical and natural language concurrently, without requiring expensive pre-training on single domains or task-specific models. 
    Interestingly, sharing weights across domains remarkably improves our model when benchmarked against state-of-the-art baselines on single-domain and cross-domain tasks.
    In particular, sharing information across domains and tasks gives rise to large improvements in cross-domain tasks, the magnitude of which increase with scale, as measured by more than a dozen of relevant metrics. 
    Our work suggests that such models can robustly and efficiently accelerate discovery in physical sciences by superseding problem-specific fine-tuning and enhancing human-model interactions.
\end{abstract}
\section{Introduction}
\label{section:introduction}
The transformer architecture~\citep{vaswani2017attention} has had a significant impact on several fields within computer science, such as language understanding~\citep{devlin2018bert}, text generation~\citep{radford2019language, brown2020language}, image understanding~\citep{dosovitskiy2020image}, multi-modal generation~\citep{ramesh2022hierarchical, saharia2022photorealistic}, among others.
Scaling language models using this architecture has proven to be a powerful and general strategy for improving generalization. This has led to the emergence of multi-task~\citep{radford2019language} and few-shot~\citep{brown2020language, winata2021language} models leveraging scale and compute~\citep{sanh2021multitask, raffel2020exploring}.

\begin{figure}[ht]
    \centering
    \includegraphics[width=\columnwidth, keepaspectratio]{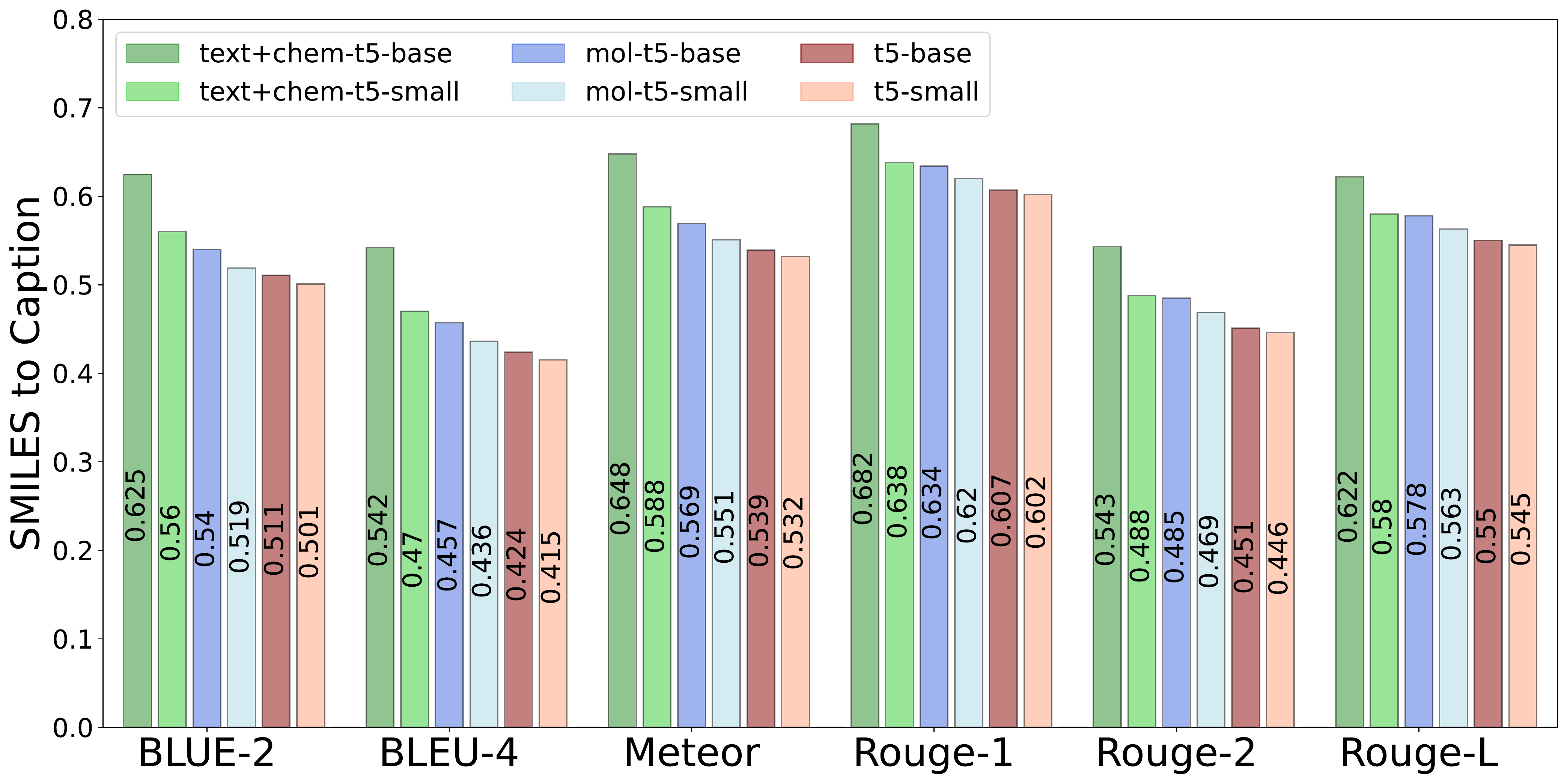}
    \caption{\textbf{Molecule to Caption task}.
    This plot compares the performance of three different models with different sizes (\emph{Text+Chem T5}-base, \emph{Text+Chem T5}-small, MolT5-base, MolT5-small, T5-base, and T5-small) on the task of converting SMILES to captions, using six different metrics: BLUE-2, BLEU-4, Rouge-1, Rouge-2, Rouge-L, and Meteor. The models are compared by plotting their scores on the y-axis. The graph shows that our proposal, \emph{Text+Chem T5}, performs the best on all metrics and improves with size, corroborating our hypothesis that joint learning on molecular and textual domains leveraging multitask learning is a powerful paradigm to bridge the gap between domains.}
    \label{fig:mol2text_figure1}
\end{figure}
\begin{figure}[ht]
    \centering
    \includegraphics[width=\columnwidth, keepaspectratio]{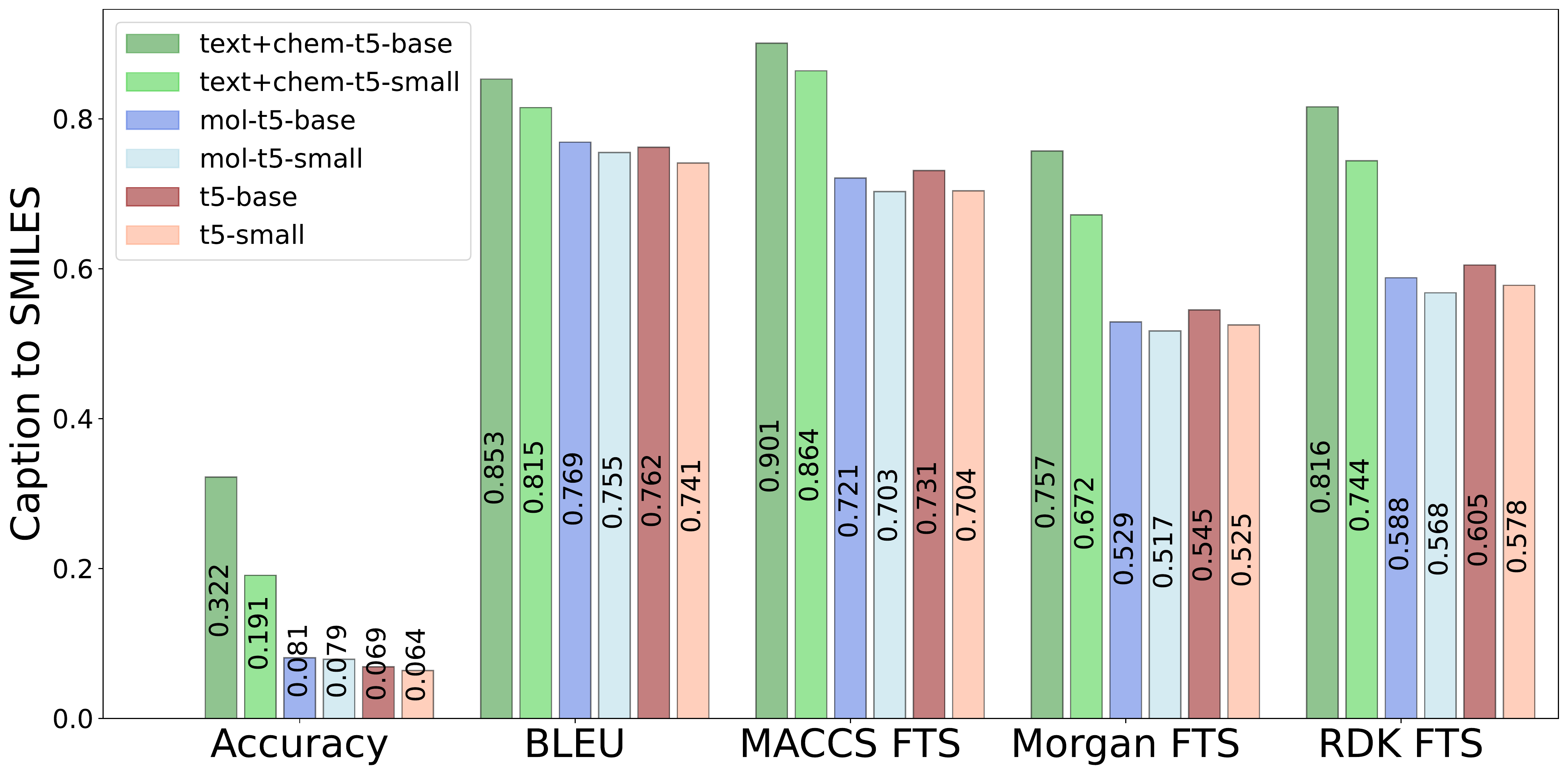}
    \caption{\textbf{Description to Molecule task}.
    This plot compares the performance of three different models with different sizes (\emph{Text+Chem T5}-base, \emph{Text+Chem T5}-small, MolT5-base, MolT5-small, T5-base, and T5-small) on the task of converting captions to SMILES, using five different metrics: Accuracy, Morgan FTS, RDK FTS, BLEU, MACCS FTS. The models are compared by plotting their scores on the y-axis. The graph shows that our proposal, \emph{Text+Chem T5}, performs the best on all metrics and improves with size, corroborating our hypothesis that joint learning on molecular and textual domains leveraging multi-task learning is a powerful paradigm to bridge the gap between domains.}
    \label{fig:text2mol_figure1}
\end{figure}

\begin{figure*}[htb]
    \centering
    \includegraphics[width=1.0\textwidth]{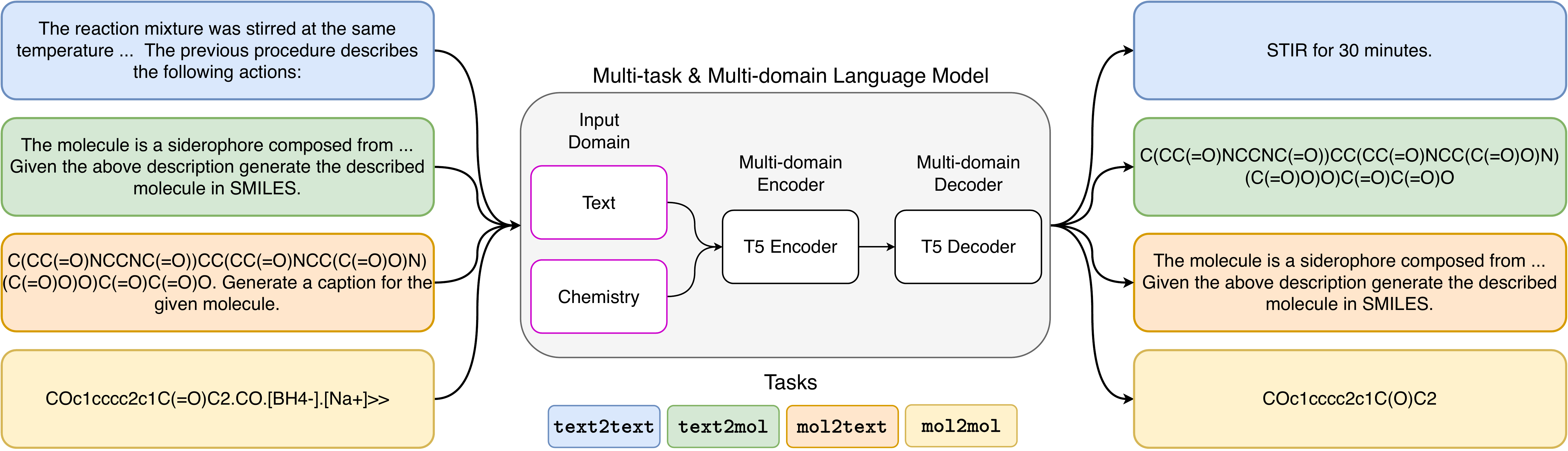}
    \caption{\textbf{\emph{Text+Chem T5} pipeline}. The \emph{Text+Chem T5} pipeline is a multi-task, multi-domain language model that integrates natural and chemical language. The model can solve language tasks, chemical tasks, and cross-domain tasks, without the need for task-specific fine-tuning or retraining. The chemical tasks that the model can solve are forward reaction prediction and retro-synthesis. The forward reaction task is about predicting the outcome of a chemical reaction based on the starting materials, and the retro-synthesis task is about predicting the starting materials required to synthesize a given chemical compound. 
    The cross-domain tasks that the model can solve are text-to-molecule (text-conditional de novo generation) and molecule-to-text (molecular captioning). The text-to-molecule task is where the model takes a textual description of a molecule as an input and generates its SMILES representation. The molecule-to-text task is where the model takes a molecule represented as SMILES and generates its human-readable textual description. 
    For the mono-domain, language task, we focus on paragraph-to-action, given a paragraph describing how to build a molecule, and output the actions required to obtain that result. The model leverages large, pre-trained single-domain models, such as T5~\citep{raffel2020exploring}, to solve all these tasks effectively. 
    The pre-trained models serve as a good starting point for fine-tuning the target distribution of tasks. 
    Further variants of the Text+Chem T5 model that were explored in this work are shown in~\autoref{fig:clm-family}.
    }
    \label{fig:clm-pipeline}
\end{figure*}

Recent developments in language models have fueled applications in engineering and science.
One notable area of success is chemistry, where ideas from natural language have been used to make significant advancements in reaction prediction~\citep{schwaller2019molecular}, conditional compound generation~\citep{born2021paccmannrl,born2021data}, retrosynthesis~\citep{schwaller2020predicting}, text-conditional de novo generation~\citep{edwards2021text2mol}, property-driven molecular design~\citep{born2022regression}, protein structure prediction~\citep{jumper2021highly}, among others.
By interpreting chemistry as a programmable language for life sciences, transformer-based models are revolutionizing the chemical discovery pipeline, significantly speeding up laboratory and design automation~\citep{o2021ai,vaucher2020automated}, and paving the way for an age of accelerated discovery in science and engineering~\citep{manica2022gt4sd}.

Despite these successes, language model advancements in the chemical domain are still limited. Specialized models must be built for each task of interest, which is time-consuming and requires a significant amount of human expertise. When multiple domains are considered, e.g., generating a novel molecule from its technical description in natural language, merging information is challenging due to the domain shift between language and chemistry. Current solutions often involve pre-training the model on large, single-domain datasets and fine-tuning on each task~\citep{edwards2021text2mol}, resulting in high computational expense, sample inefficiency, and the need to repeat this process for each use-case.

In light of these limitations, it is worth considering the feasibility of a more efficient and general multi-task model that can translate between the textual and chemical domains. This type of model would be particularly useful in cases where large amounts of data are not available and domains are unbalanced. Such models would also be critically important for tasks where information sharing is essential, like molecular captioning (given a molecule, describe it in natural language) or text-conditional de-novo generation (given a description, generate a molecule).

In this work, we propose a multi-task transformer for natural language and chemical translation, 
\emph{Multitask Text and Chemistry T5} (\emph{Text+Chem T5} for brevity). We focus on transfer learning in the chemical domain, with a specific emphasis on cross-domain tasks, tasks that involve chemistry and natural language concurrently. 
The performance comparisons of \emph{Text+Chem T5} to previous models in \autoref{fig:mol2text_figure1} and~\autoref{fig:text2mol_figure1} show the consistent superiority of our approach for cross-domain tasks across various NLP-based evaluation metrics, like BLEU~\cite{papineni2002bleu}, Rouge~\cite{lin2004rouge} and Meteor~\cite{banerjee-lavie-2005-meteor}.
\emph{Text+Chem T5} does not rely on expensive mono-domain pre-training, task-specific fine-tuning, or separate heads for each task. Our model can be directly used on a variety of chemical and natural language-based tasks in a mono-domain and cross-domain setup (cf.~\autoref{fig:clm-pipeline} for a graphical abstract).
Notably, \emph{Text+Chem T5} enables the execution of complex molecular discovery workflows \textit{with a single model}, a previously unreported ability, that, as we demonstrate, is even beyond the capability of foundation models like ChatGPT or Galactica~\citep{taylor2022galactica}.

\textbf{Contribution.}
Our work presents the following key contributions: 
\begin{itemize}
    \item[\textbf{(i)}] We introduce a novel cross-domain, multi-task chemical language model (\emph{Multitask Text and Chemistry T5}) that effectively bridges natural and chemical languages by enabling translation between the domains. 
\item[\textbf{(ii)}] We propose an efficient training strategy that leverages the strengths of both single-domain and multi-domain tasks to adapt single-domain models for cross-domain tasks. This eliminates the need for costly pre-training on large mono-domain datasets and task-specific fine-tuning, while at the same time improving cross-domain translation by sharing information between tasks and across domains. 
\item[\textbf{(iii)}] We provide experimental validation on benchmark datasets for single and cross-domain tasks, demonstrating that our model is competitive with state-of-the-art methods specialized for single tasks. We also conduct a thorough analysis of various modeling choices, including encoder-decoder architecture, the use of frozen vs. learnable encoders, and single-domain vs. multi-domain encoders. 
\end{itemize}
\section{Background}
\label{section:background}
Our model is designed to handle tasks that span multiple domains (see Fig.~\ref{fig:clm-pipeline}), specifically chemistry-based tasks (\texttt{mol2mol}), textual-based tasks (\texttt{text2text}), and cross-domain tasks (\texttt{mol2text} and \texttt{text2mol}).

In this section, we present an overview of recent advances in generative language models for transfer and multi-task learning in the natural language and chemical domains. We examine the limitations of current models, particularly in cross-domain generation, and demonstrate the necessity for our proposed approach.
\begin{table}[ht]
\setlength{\tabcolsep}{3pt}
\begin{center}
 \caption{\textbf{Language Models for Chemistry}.
We compare language models by expressivity and flexibility. 
The column w/ text pretrain (with text pretraining) indicates if a model is pretrained on text before finetuning.
The column w/ chem pretrain (with chemistry pretraining) indicates if a model is pretrained on chemistry before finetuning.
In particular, if a model pre-trained on one domain (text for T5, MD-T5, MolT5, Text+Chem T5, and chemistry for T5Chem) is also pre-trained on the second domain (chemistry for T5, MD-T5, MolT5, Text+Chem T5 and text for T5Chem)  before fine-tuning on the cross-domain tasks, we consider such model pre-trained on both domains (like MolT5) before fine-tuning. 
In general, more domains are used for pretraining, and more domain-specific data and computational resources are needed. So having a model that is pre-trained on only one domain is preferable.
The column "encoder-sharing" indicates whether the model shares encoders between domains or not.
The table shows that \emph{Text+Chem T5} is the only model that does not leverage expensive pre-training on both domains, and at the same time leverages multi-tasking, multi-domain learning, and encoder sharing, making \emph{Text+Chem T5} more expressive and feature-rich than the other models in the literature. 
$e^2$: the model uses domain-specific encoders. MD: multi-domain. MT: multi-task.}
\resizebox{0.99\columnwidth}{!}{
		\begin{tabular}{l | ccccc}
		\toprule
		 & w/ text & w/ chem & multi- & multi- & encoder \\
		 & pretrain & pretrain & task & domain & sharing \\
		\midrule
		T5~\citep{raffel2020exploring}     & \cmark  & \xmark & \cmark & \xmark & \cmark       \\
		MD-T5 (finetuned)  & \cmark  & \xmark & \xmark & \cmark & \cmark       \\
		T5Chem~\citep{lu2022unified} & \xmark  & \cmark & \cmark & \xmark & \cmark       \\
		MolT5~\citep{edwards2022translation}  & \cmark & \cmark & \xmark & \cmark & \xmark       \\
		\midrule
            MDMT$e^2$-CLM (ours) & \cmark & \cmark & \cmark & \cmark & \xmark \\
		\textbf{Text+Chem T5} (ours) & \cmark & \xmark & \cmark & \cmark & \cmark      \\
		\bottomrule
		\end{tabular}
  }
	\label{tab:multi}
	\end{center}
\end{table}
Transformers, as presented in~\citep{vaswani2017attention}, are widely used in language modeling. T5~\citep{raffel2020exploring}, a transformer trained on a diverse set of tasks, has demonstrated impressive generalization and adaptation capabilities in multi-tasking and multi-modal generation~\citep{saharia2022photorealistic}. We use T5 as the backbone of our work.

Specialized models for chemistry have been developed, such as Molecular Transformers~\citep{schwaller2019molecular} and the RXN family~\citep{schwaller2018found, schwaller2020predicting, toniato2021unassisted, vaucher2020automated}, which address tasks like forward reaction prediction and molecular retrosynthesis. However, these models require separate models for each task, leading to increased computational cost and a need for specialized expertise for each task.

T5Chem~\citep{lu2022unified} offers a unified multi-tasking model for chemistry, using a single model for tasks like reaction prediction, regression, and classification. However, T5Chem relies on task-specific heads, is restricted to the chemical domain and has limited applicability in different sub-domains.

MolT5~\citep{edwards2022translation} addresses the difficult problem of cross-domain generation by linking natural language and chemistry, tackling tasks such as text-conditional de novo molecule generation and molecule captioning. However, it relies on costly pre-training on large mono-modality datasets, as well as per-task fine-tuning for each multi-modal task, which in turn limits its ability to leverage the multi-tasking capabilities of T5 and the sharing of information between tasks.

Despite the progress made using multi-task learning in natural language processing, transferring these advancements to the chemical domain remains a challenge. Current models focus on optimizing specific tasks~\citep{schwaller2019molecular}, multi-tasking learning~\citep{lu2022unified} or translation between text and chemistry~\citep{edwards2022translation}, but still struggle with handling multiple tasks across domains without incurring in the expense of large, sample-inefficient pre-training and fine-tuning. 
While multi-tasking within a single domain may be feasible, multi-tasking across multiple domains is still a challenge. Our proposed solution is a simple training strategy, which utilizes pre-trained transformers for each modality and a learnable, small output decoder to merge modalities in the later stages of the model, thus addressing these challenges.
\section{Method}
\label{section:method}
\begin{figure*}[!ht]
    \centering
    \includegraphics[width=1.0\textwidth, keepaspectratio]{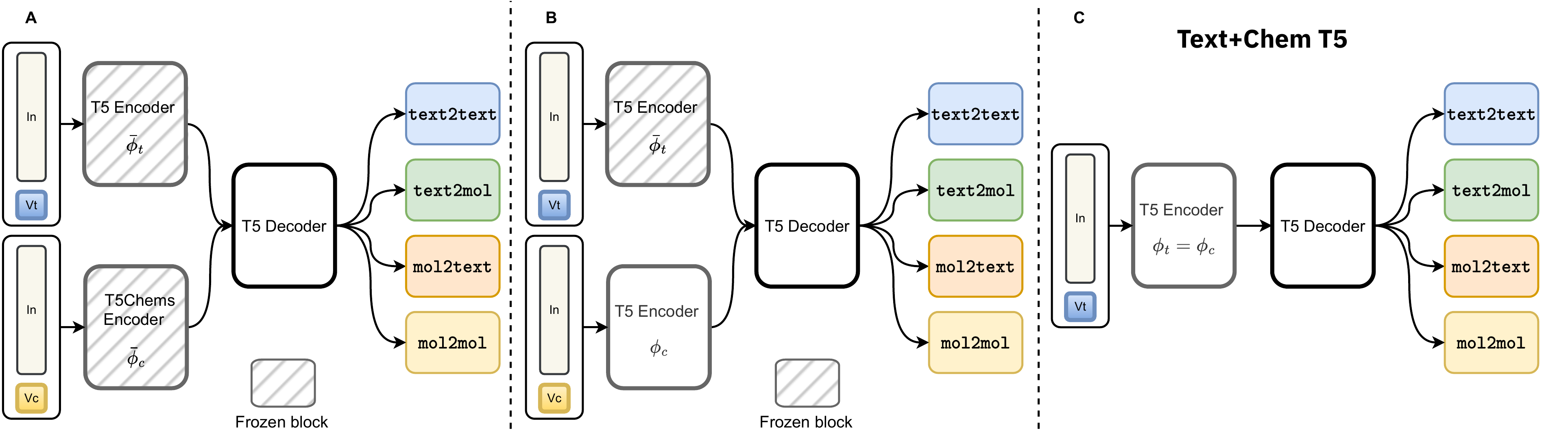}
    \caption{\textbf{The Chemical Language Model (CLM) family}.
    The caption describes three different approaches to building a multi-domain model for text and chemistry tasks. 
    \underline{\textbf{A}}: a multi-domain model is built without the need to retrain the single-domain encoders (no enc-sharing, no enc-training). Instead, two frozen sets of weights ($\bar \phi_t$, $\bar \phi_c$) are used for the text and chemistry encoders respectively. These weights are extracted from large, pre-trained language encoders, such as T5~\citep{raffel2020exploring} and T5Chem~\citep{lu2022unified}.
    \underline{\textbf{B}}: a multi-domain model is still built using two sets of weights. However, the chemistry encoder is fine-tuned (enc-training) while the text encoder remains frozen (no enc-sharing). The fine-tuning process starts from a pre-trained T5 checkpoint(1.0) fine-tuned on chemistry data. 
    \underline{\textbf{C}}: 
    The final, proposed \emph{Text+Chem T5} model.
    the encoders are merged, using a joint encoder for text and chemistry ($\phi_t = \phi_c$) and trained jointly on the multi-domain and multi-task data (enc-training, enc-sharing). This approach allows the model to be fine-tuned on a variety of tasks and domains, which improves its generalization capabilities. A T5 decoder is used and no separate heads are used for each task or domain. 
    The sharing of information between tasks and domains enriches the model's generalization.
    $V_t$ is the vocabulary for text and $V_c$ is the one for chemistry.
    }
    \label{fig:clm-family}
\end{figure*}

\paragraph{Model.}
Our goal is to develop a multi-task, multi-domain model for natural and chemical language. 
To achieve this, we use a T5 backbone, an encoder-decoder transformer architecture specifically proposed for multi-tasking~\citep{raffel2020exploring}. 
The encoder-decoder architecture is especially suited for cross-domain tasks because we can explore a family of architectural choices modifying the encoder without the need to modify the decoder (see Fig.~\ref{fig:clm-family}). Doing so we can ablate our multi-domain multi-task model with variations where we use two encoders (one for each domain), we consider a different model as chemistry encoder, and we freeze the encoders (more details in Table~\ref{tab:experiments-ablation-clm}).
We name our model \emph{Text+Chem T5}.%

\paragraph{Tasks Distribution.}
The \emph{Text+Chem T5}  model is designed to handle tasks that span multiple domains, specifically chemistry-based tasks (\texttt{mol2mol}), textual-based tasks (\texttt{text2text}), and cross-domain tasks (\texttt{mol2text} and \texttt{text2mol}).
Our objective is to train the model to learn a mapping between languages without losing proficiency in the original languages, similar to cross-domain generative tasks in the context of language translation.
To achieve this, we follow the task-prompting method outlined in~\citet{raffel2020exploring}. Specifically, we focus on the following tasks for each domain:

$\bullet$ \texttt{mol2mol} is a mono-domain task that is focused on chemical reactions, it has two sub-tasks:

\underline{Forward reaction}. Given precursors (optionally including reagents and/or enzymes), the task is to generate the main product of the reulting chemical reaction. This sub-task is a classic example of a forward reaction prediction task, the model has to predict the outcome of a chemical reaction based on the starting chemicals.

\underline{Retrosynthesis}. Given the product of a chemical reaction, the goal is to find the precursors (optionally including reagents and/or enzymes). This sub-task is an example of a retrosynthesis task, which is the inverse of a forward reaction prediction task. The model needs to predict the starting chemicals that would be required to synthesize a given compound.
Note that we consider one-step retrosynthesis only.

$\bullet$ \texttt{mol2text} is a cross-domain task that is focused on generating natural text from chemical input.

\underline{Molecular captioning}. Given a molecule represented as SMILES (Simplified Molecular Input Line Entry System), the goal is to generate a textual description of such molecule. This task is an example of a cross-domain task as it involves both chemistry and natural language processing. The model is to generate a human-readable description of a chemical compound based on its SMILES representation.

$\bullet$ \texttt{text2mol} is a cross-domain task that is focused on generating chemical representation from text.

\underline{Text-conditional de novo generation}. In this cross-domain task, a textual paragraph describing the molecule is provided and the goal is to output a SMILES representation for such molecule. This task is an example of a cross-domain task as it involves both natural language processing and chemistry. The model should generate the SMILES representation for a chemical compound based on its textual description.

$\bullet$ \texttt{text2text} is a mono-domain task that is focused on natural language processing.

\underline{Paragraph to action}. The task is to generate the action sequence for a certain chemical reaction described in natural language. The model is to take a natural language description of a chemical reaction and generates a step-wise execution protocol to carry out that reaction. This task is an example of text generation in natural language processing, and it's focused on understanding the chemical reactions and converting them into a list of actions.

\paragraph{Merging Domains.}
As shown in Fig.~\ref{fig:clm-family} and Table~\ref{tab:experiments-ablation-clm}, we ablate \emph{Text+Chem T5}  architecture using different encoder setups. In particular, we want to explore the cross-domain performance when using a different encoder for each domain. In this scenario, we need a mechanism to aggregate information from the natural language and chemistry encoder. 
Given the difference between domains, it is crucial to find an expressive way to merge information at the late stage in input to the decoder.
One way to accomplish this is to simply average the encoder output embeddings.
However, a more expressive approach is to use a cross-attention approach, loosely inspired by the contextual attention in~\citet{born2021data}. Specifically, by selecting one domain as the base domain, we can translate information in a powerful way. This approach is also well suited to scale to multiple domains.
We denote $H_t \in \bR^{(n_t, h_t)}$ as the output of the base domain encoder, where $n_t$ is the sequence length (number of tokens) and $h_t$ is the hidden dimensionality. We also consider a second domain, $H_m \in \bR^{(n_m, h_m)}$, where $n_m$ is the number of tokens and $h_m$ is the dimensionality for the adaptation domain (e.g. chemistry information).
We merge this information using cross-attention by setting the base domain as the queries 
$Q = f_t(H_t) \in \bR^{(n_t, d)}$, and the adaptation domain as the keys 
$K = f_k(H_m) \in \bR^{(n_m, d)}$ and values $V = f_v(H_m) \in \bR^{(n_m, d)}$.
In practice, we compute $W = \sigma(Q, K^T) \in \bR^{(n_t, n_m)}$ and finally $H_{tm} = W,V \in \bR^{(n_t, d)}$.
We can then apply this block to $H_{tm}$ in a hierarchical fashion, setting $H_t = H_{tm}$. Finally, the output of this attention network is fed to the T5 decoder. The use of text as the base domain means that we can feed the final $H_t$ directly to the T5 decoder, without the need for additional adaptation of the architecture.
An alternative and more expressive approach is to merge the adaptation mode into the base mode to obtain $H_{tm}$, and then merge the base mode into the adaptation mode to obtain $H_{mt}$, and combining these intermediate results to obtain the representation input for the decoder $H'_{tm}$. 

\section{Experiments}
\label{section:experiment}
\paragraph{Setup.}
We evaluate the model's performance on five tasks: forward and backward reaction prediction in chemistry, text-conditional de novo molecule generation and molecule captioning across domains, and paragraph-to-action in the language domain.
The training process is carried out using the language modeling trainer based on Hugging Face transformers~\citep{Wolf_Transformers_State-of-the-Art_Natural_2020} and PyTorch Lightning~\citep{Falcon_PyTorch_Lightning_2019} from the GT4SD library~\citep{manica2022gt4sd}.
To initialize our transformer model, we choose to use the natural language domain, as it has the most available data. For this reason, we use T5-small and T5-base as pretrained bases for our respective models.
Details on the models' hyperparameters can be found in Appendix~\ref{app:details}.

\paragraph{Dataset.}
To train our model, we generated a multi-domain and a multi-task dataset by aggregating available datasets for each task of interest. Specifically, we leveraged the dataset used in \citet{toniato2021unassisted} which has been derived by Pistachio dataset~\citep{pistachio} (release of 18 November 2019) for \texttt{mol2mol} tasks. This dataset contains 2.3M reactants-products pairs as training set, 10k pairs as validation set and 10k pairs as testing set. For the paragraph-to-actions task, we relied on the procedures dataset (2.16M samples in the training set and 270k samples in the validation set and in the testing set) presented in~\citeauthor{vaucher2020automated}. Finally, we use the CheBI-20 dataset~\citep{edwards2021text2mol,edwards2022translation} ($\sim$26k molecule-description pairs as training set, $\sim$3k pairs as validation set and $\sim$3k as testing set) for the description-to-smiles and smiles-to-caption tasks. The final training dataset is balanced between tasks by having 2.3M samples of each task. This leads to a training set of total 11.5M samples. For the task in which their training sets contain fewer samples, we augmented them by repeating the existing samples more than one time. A second augmented version of the training set was also been constructed by including further reactants-products pairs. This second version had 6.7M reaction pairs and in total 33.5M samples. For the second augmented dataset, we again followed an \emph{equal mixing} strategy~\cite{raffel2020exploring}, i.e., a balance in the number of instances among tasks was ensured. The use of the augmented dataset in the presented results is indicated by the prefix `augm` in the respective table rows. In the rest cases, the first version of the multi-task dataset has been used to train the respective model. In both datasets, we rely on prompts for the task definition. The prompt template that has been used can be found in appendix (see~\autoref{app:prompts}).

\begin{table*}[ht]
    \centering
    \caption{\textbf{Results across domains and tasks}.
We evaluate T5, a multi-task model for the textual domain, finetuned for each class of tasks; for the chemical domain, we consider also specialized models for forward and retrosynthesis (RXN family); and MolT5, a multi-domain model for the textual and chemical domains.
Tasks evaluated include chemical-based tasks (forward and retrosynthesis), cross-domain tasks (text-conditional de novo generation and molecule captioning), and textual-based tasks (paragraph to action). Our goal is to leverage multi-task learning to improve cross-domain translation between chemistry and text.
The forward and retrosynthesis RXN baseline results are re-evaluations of the original models~\citep{schwaller2019molecular, schwaller2020predicting} as presented in~\citet{toniato2021unassisted}. For the forward prediction task the metric is accuracy; for the retrosynthesis task the metric is roundtrip accuracy~\cite{schwaller2020predicting}; for all the other tasks the BLEU score. For more metrics see Tables~\ref{tab:experiments-smiles-text} and \ref{tab:experiments-text-smiles} and Table~\ref{tab:model-size} for model sizes.
    }
    \resizebox{0.99\textwidth}{!}{
    \begin{tabular}{cc | ccccc}
    \toprule
    Domain & &\multicolumn{2}{c}{\texttt{mol2mol}} & \multicolumn{2}{c}{cross-domain}& \multicolumn{1}{c}{\texttt{text2text}} \\ 
    Task & Size & forward & retrosynthesis & \texttt{text2mol} & \texttt{mol2text} & paragraph-actions \\
    \midrule
    T5 (fine-tuned)~\citep{raffel2020exploring} & small & 0.603 & 0.245  & 0.499  & 0.501 & \textbf{0.953} \\
   T5 (fine-tuned)~\citep{raffel2020exploring} & base & 0.629 & -  & 0.762  & 0.511  & -  \\
    RXN-forward~\citep{toniato2021unassisted} & - & \textbf{0.685} &  -  & - & - & -\\
    RXN-retrosynthesis~\citep{toniato2021unassisted} & - & - &  \textbf{0.733}  & - & - & -\\
    RXN-paragraph2actions~\citep{vaucher2020automated} & - & - &  -  & - & - & 0.850\\
    MolT5~\citep{edwards2022translation} & small & -  & - & 0.755 & 0.519 & - \\
    MolT5~\citep{edwards2022translation}  & base  & -  & - & 0.769 & 0.540 & - \\
    \midrule
    \textbf{\emph{Text+Chem T5}  (ours)}             & small & 0.412  & 0.249  & 0.815 & 0.553 & 0.929 \\
    \textbf{\emph{Text+Chem T5}  (ours)}              & base  & 0.459  & 0.478  & 0.750 & 0.580 & 0.935 \\
    \textbf{\emph{Text+Chem T5-augm} (ours)}             & small & 0.413  & 0.405  & 0.815 & 0.560 & 0.926 \\
    \textbf{\emph{Text+Chem T5-augm} (ours)}              & base  & 0.594  & 0.372  & \textbf{0.853} & \textbf{0.625} & 0.943 \\
    \bottomrule
    \end{tabular}
    }
    \label{tab:experiments-main}
\end{table*}

\begin{table*}[ht]
\small
    \centering
    \caption{\textbf{Results of the SMILES to Caption ($\texttt{mol2text}$) task}. 
    The baselines include Transformer~\citep{edwards2022translation}, T5 (fine-tuned), and MolT5~\citep{edwards2022translation}. The metrics used in the table include BLEU-2, BLEU-4, Rouge-1, Rouge-2, Rouge-L, and Meteor, all of which are common metrics used to evaluate text generation models. The table shows that our proposed model, Text+Chem T5, outperforms the other baselines in all the metrics. Overall, \emph{Text+Chem T5}  is able to generate more accurate and informative captions for SMILES.
    }
    \resizebox{0.99\textwidth}{!}{
    \begin{tabular}{c c | cccccc}
    \toprule
     & Size & BLEU-2 $\uparrow$ & 
     BLEU-4 $\uparrow$ & Rouge-1 $\uparrow$ & Rouge-2 $\uparrow$ &Rouge-L $\uparrow$ &Meteor $\uparrow$\\
    \midrule
    Transformer~\citep{edwards2022translation} & - &0.061&0.027&0.188&0.0597&0.165&0.126\\
    T5 (fine-tuned)~\citep{raffel2020exploring} & small &0.501&0.415&0.602&0.446&0.545&0.532\\
    MolT5~\citep{edwards2022translation} & small & 0.519 & 0.436 & 0.620 & 0.469 & 0.563 & 0.551\\
    \textbf{\emph{Text+Chem T5}  (ours)} & small & 0.553 &0.462 & 0.633&0.481&0.574&0.583\\
    \textbf{\emph{Text+Chem T5-augm} (ours)} & small &\textbf{0.560}&\textbf{0.470}& \textbf{0.638}&\textbf{0.488}&\textbf{0.580}&\textbf{0.588}\\
    \midrule
    T5(fine-tuned)~\citep{raffel2020exploring} & base &0.511 & 0.424 & 0.607 & 0.451 & 0.550 & 0.539\\
    MolT5~\citep{edwards2022translation} & base &0.540 & 0.457 & 0.634 & 0.485 & 0.578 & 0.569\\
    \textbf{\emph{Text+Chem T5}  (ours)} & base & 0.580 & 0.490 & 0.647 & 0.498 & 0.586 & 0.604 \\
    \textbf{\emph{Text+Chem T5-augm} (ours)} & base & \textbf{0.625} & \textbf{0.542} & \textbf{0.682} & \textbf{0.543} & \textbf{0.622} & \textbf{0.648} \\
    \bottomrule
    \end{tabular}
    }
    \label{tab:experiments-smiles-text}
\end{table*}

\begin{table*}[ht]
\setlength\tabcolsep{4.0pt}
    \centering
    \caption{\textbf{Results of the Description to SMILES (\texttt{text2mol}) task}. 
    The performance of the models is evaluated by BLEU score, Accuracy, Levenshtein distance, and additional metrics (see Evaluation). The results show that the proposed model (Text+Chem T5) outperforms other baselines in all metrics. These results demonstrate the effectiveness of the proposed model in translating from natural language to SMILES representation of molecules.
    }
    \resizebox{0.99\textwidth}{!}{
    \begin{tabular}{cc | cccccccc}
    \toprule
     & Size & BLEU score $\uparrow$ &Accuracy $\uparrow$&Levenshtein $\downarrow$ &MACCS FTS$\uparrow$ &RDK FTS$\uparrow$ &Morgan FTS$\uparrow$ &FCD$\downarrow$ &Validity$\uparrow$ \\
    \midrule
    Transformer~\citep{edwards2022translation} & - & 0.499 & 0 & 57.66 & 0.480 & 0.320 & 0.217 & 11.32 & 0.906\\
    T5 (fine-tuned)~\citep{raffel2020exploring} & small &0.741 & 0.064 & 27.7 & 0.704 & 0.578 & 0.525 & 2.89  & 0.608 \\
    MolT5~\citep{edwards2022translation} & small & 0.755 & 0.079 & 25.99 & 0.703 & 0.568 & 0.517 & 2.49 & 0.721 \\
    \textbf{\emph{Text+Chem T5}  (ours)} & small &0.739 & 0.157 &28.54 & 0.859 & 0.736 & 0.660 & 0.066 & 0.776 \\
    \textbf{\emph{Text+Chem T5-augm} (ours)} & small &\textbf{0.815} &\textbf{0.191} &\textbf{21.78} & \textbf{0.864} & \textbf{0.744} & \textbf{0.672} & \textbf{0.060} & \textbf{0.951} \\
    \midrule
    T5 (fine-tuned)~\citep{raffel2020exploring} & base & 0.762 & 0.069 & 24.95 &0.731 & 0.605 & 0.545 & 2.48 & 0.660  \\
    MolT5~\citep{edwards2022translation} & base & 0.769 & 0.081 & 24.49 & 0.721 & 0.588 & 0.529 & 0.218 & 0.772\\
    \textbf{\emph{Text+Chem T5}  (ours)} & base & 0.750 & 0.212 & 27.39 & 0.874 & 0.767 & 0.697 & 0.061 & 0.792 \\
    \textbf{\emph{Text+Chem T5-augm} (ours)} & base & \textbf{0.853} & \textbf{0.322} & \textbf{16.87} & \textbf{0.901} & \textbf{0.816} & \textbf{0.757} & \textbf{0.050} & \textbf{0.943} \\
    \bottomrule
    \end{tabular}
    }
    \label{tab:experiments-text-smiles}
\end{table*}

\paragraph{Evaluation.}
Evaluating the model is challenging as it spans multiple domains. For this reason, we treat each task separately and we rely on a combination of NLP based as well as task-specific metrics.
Moreover, we provide qualitative evaluations as well as comparisons to broader multitask models like ChatGPT and Galactica (see Appendix~\ref{appx:comparison-workflow}).

For the molecule-to-text task (\texttt{mol2text}), we consider the following metrics:
BLEU-2 and BLEU-4~\citep{papineni2002bleu} are metrics used to evaluate the quality of machine-generated text by comparing it to a reference text. BLEU-2 computes the 2-grams (word bigrams) overlap between the generated text and reference text, and BLEU-4 extends this to 4-grams. A higher score on the BLEU metric indicates a higher level of similarity between the generated text and reference text.
ROUGE-1, ROUGE-2, and ROUGE-L~\citep{lin2004rouge} are similar to BLEU, but compute the recall-overlap of unigrams, bigrams, and longest common subsequences between the generated and reference texts.
METEOR~\citep{banerjee-lavie-2005-meteor} is a metric that uses a combination of unigram precision, recall, and a synonym-matching component to evaluate the generated text against the reference text.
It's designed to be more sensitive to fluency, meaning, and structure than BLEU.

For the text-to-molecule task (\texttt{text2mol}), we consider the following metrics:
BLEU scores, Accuracy, Levenshtein distance, MACCS-FTS~\citep{durant2002reoptimization}, RDK-FTS~\citep{tanimoto1958}, Morgan-FTS~\citep{rogers2010extended} and FCD~\citep{preuer2018frechet}. Furthermore, we report validity as the percent of molecules which can be processed by RDKit as in~\citet{ edwards2022translation}.
For this task, the accuracy is the number of correctly generated molecules made by the model divided by the total number of samples.
Levenshtein distance is a string similarity metric counting the number of edits (insertions, deletions, or substitutions) required to change one sequence into the other~\citep{levenshtein1966binary}. 
It is often used in natural language processing for tasks such as spell-checking and speech recognition.

\paragraph{Baselines.}
We compare our proposed method with several baselines including a standard Transformer model, T5 fine-tuned on each task, the RXN family, and MolT5. For T5, MolT5, and our model, we consider a small (40M) and a base (220M) version.

\subsection{Results} 
With these experiments, we aim to demonstrate: 
\textbf{(i)} the effectiveness of a joint, cross-domain multi-task language model in improving generalization on cross-domain tasks; 
\textbf{(ii)} that by leveraging pre-trained single-domain information, we can avoid the need for costly pretraining and task-specific fine-tuning; 
\textbf{(iii)} that by sharing information not only between tasks and domains but also between encoder weights, we can achieve the best cross-domain translation. Lastly, we will illustrate how our approach benefits from increased scale, providing a general paradigm for language models in the scientific domain.

\paragraph{Cross-domain Tasks.}
Table~\ref{tab:experiments-main} presents the results of different models that use multi-task learning for different domains and tasks. The models are T5, a multi-task model for the textual domain, and MolT5, a multi-domain model for the textual and chemical domains. 
The generalization capabilities of these models are evaluated on tasks from both the textual and chemical domains, including chemical-based tasks (forward and retrosynthesis), cross-domain tasks (text-conditional de novo generation and molecule captioning), and textual-based tasks (text generation). The table includes metrics for each task and model, with the best results highlighted in bold. \textit{Text+Chem T5} is the proposed multi-domain and multi-task model.
We also compare with additional baselines such as transformer, T5, RXN-family models and MolT5.

\paragraph{Molecule to Text.}
Table~\ref{tab:experiments-smiles-text} presents the results of different models evaluated on the SMILES to Caption (mol2text) task. The models include Transformer, T5, MolT5, and \textit{Text+Chem T5}. The performance of each model is evaluated using several metrics, including BLEU-2, BLEU-4, Rouge-1, Rouge-2, Rouge-L, and Meteor. The best results are highlighted in bold.
\emph{Text+Chem T5}  model performs best on all evaluation metrics compared to the other models. Specifically, it achieved the best score on BLEU-2, BLEU-4, Rouge-1, Rouge-2, Rouge-L, and Meteor. The BLEU scores indicate that this model generated captions that were highly similar to the reference captions and the Rouge score indicates that this model is generating grammatically correct and fluent text. The Meteor score indicates this is the model that generated the most similar sentence to reference.
\emph{Text+Chem T5}  model was able to achieve a BLEU-2 and BLEU-4 scores of 0.625 and 0.542, respectively. These are relatively high scores and indicate that this model is able to generate captions that are highly similar to the reference captions. The Rouge-1 score of 0.647 and Rouge-2 score of 0.498 similarly indicate that this model is generating captions that are grammatically correct and fluent. The Rouge-L score of 0.586 and Meteor score of 0.604 also indicate this model is generating high-quality captions.

\paragraph{Text to Molecule.}
Table~\ref{tab:experiments-text-smiles} presents the results of different models evaluated on the Caption to SMILES (text2mol) task. The models include Transformer, T5, MolT5, and \textit{Text+Chem T5}. The performance of each model is evaluated using BLEU score, Accuracy, and Levenshtein distance. The best results are highlighted in bold.
\emph{Text+Chem T5}  performed best among all the models in the task. Specifically, it achieved the highest BLEU score of 0.853, indicating that it generated SMILES strings that were highly similar to the reference SMILES. 
The accuracy indicates that \emph{Text+Chem T5}  generates more than 32\% of correct SMILES.
The Levenshtein distance is a measure of the similarity between two strings. A smaller Levenshtein distance indicates that the generated SMILES is more similar to the reference SMILES. The Levenshtein distance of 16.87 for the \emph{Text+Chem T5}  model indicates that this model was able to generate SMILES strings that were very similar to the reference SMILES.
Overall, the table suggests that \emph{Text+Chem T5}  model was the best among all the models for the Caption to SMILES task, as it performed well on all the metrics.

\paragraph{Architecture Ablation.}
Table~\ref{tab:experiments-ablation-clm} presents the results of an ablation study on the aggregation and encoder strategy for the cross-domain tasks of \texttt{text2mol} and \texttt{mol2text}. 
The different models considered in the ablation study include MD$e^2$-CLM, MDMT$e^2$-CLM, and \textit{Text+Chem T5}. The performance of each model is evaluated using the text2mol and mol2text metrics. The best results are highlighted in bold.
\begin{table}[ht]
\setlength\tabcolsep{3.0pt}
    \centering
    \caption{\textbf{Ablation study for different aggregation and encoder strategies for cross-domain tasks}. 
    The objective of this study is to understand how the different choices of aggregation and encoder strategies affect the performance of the model. The tasks considered are text-to-chemistry (\texttt{text2mol}) and chemistry-to-text (\texttt{mol2text}). The evaluation metrics used in the table are BLEU scores.
It can be observed that the proposed \emph{Text+Chem T5}  model outperforms the other models in both the text-to-chemistry and chemistry-to-text tasks, and thus the best approach is to use encoder sharing and tuning and not use any specific aggregation strategy. 
agg: aggregation mechanism; cross-att: cross-attention for aggregation. MD: multi-domain; MT: multi-task;
$e^2$: the model uses domain-specific encoders.}
\resizebox{0.99\columnwidth}{!}{
\begin{tabular}{c | ccc | cc}
    \toprule
    \multirow{2}{*}{Model} & \multirow{2}{*}{agg} & enc- & enc- & \texttt{text2} & \texttt{mol2} \\
    & & sharing & tuning & \texttt{mol} & \texttt{text} \\
    \midrule
    MD$e^2$-CLM       & mean      & \xmark&  \xmark   & 0.572 & 0.123 \\
    MD$e^2$-CLM       & cross-att & \xmark &  \xmark  & 0.702 & 0.274 \\
    MDMT$e^2$-CLM     & cross-att & \xmark &   \xmark & 0.247 & 0.119 \\
    MDMT$e^2$-CLM & cross-att & \xmark&  \cmark & 0.211 & 0.075 \\
    \textbf{Text+Chem T5}         & - & \cmark&  \cmark & \textbf{0.750} & \textbf{0.580} \\
    \textbf{Text+Chem T5-augm}         & - & \cmark&  \cmark & \textbf{0.853} & \textbf{0.625} \\
    \bottomrule
    \end{tabular}
    }
    \label{tab:experiments-ablation-clm}
\end{table}
The table compares different variants of the model (Figure~\ref{fig:clm-family}). MD$e^2$-CLM denotes that the model has different encoders for each domain, whereas MDMT$e^2$-CLM denotes that the model has different encoders for each domain and is trained using multiple tasks. The models are further differentiated based on the aggregation and encoder-sharing strategy. "Agg" denotes the method used for aggregating information from the different tasks, "enc-sharing" denotes whether the encoders are shared across tasks, and "enc-tuning" denotes whether the encoders are fine-tuned for each task.
\emph{Text+Chem T5}  performed the best among all the models for both the \texttt{text2mol} and \texttt{mol2text} tasks. It achieved the highest score of 0.853 on the \texttt{text2mol} task and 0.625 on the \texttt{mol2text} task.
The model that performed best is\textit{ Text+Chem T5}, it has achieved the highest scores on both cross-domain tasks, it was implemented using shared encoders and fine-tuning approach.
Using a shared encoder and fine-tuning strategy for the CLM model improves performance on cross-domain tasks. Also, the ablation study has indicated that the aggregate method didn't play a big role in the performance of the model, but shared encoders and fine-tuning approach had the most significant effect on the performance of the model.

\paragraph{Model Size.}
\emph{Text+Chem T5}  results improve by increasing the model size.
Figure~\ref{fig:mol2text_figure1} shows the trend for different metrics (x-axis) for the SMILES to Caption (\texttt{mol2text}) task. 
We report results for T5-fine-tuned, MolT5 and \emph{Text+Chem T5}  in two different model sizes: small (60M parameters) and base (220M parameters). These are standard sizes for T5-based models.
We see how the results for \emph{Text+Chem T5}  not only improve increasing the model capacity but improve much faster than baselines with similar capacity. We see similar trends in Table~\ref{tab:experiments-smiles-text} and Table~\ref{tab:experiments-text-smiles}, corroborating the idea that joint multi-task training on textual and molecular domains is an effective mechanism to enrich representations in language models and share information among tasks and domains.

\paragraph{Dataset Size.}
The augmented version of the dataset contributes to improved performance across the whole span of tasks. This improvement is especially high in the tasks in which the requested output modality is SMILES which is totally aligned with the new information that we have incorporated in the augmented version of the dataset. This observation underlines the need for a high volume of data in such training strategies to assist the model in better understanding the different domains or modalities of interest. 

\paragraph{Qualitative Evaluation.}
Finally, we conduct a qualitative assessment of \emph{Text+Chem T5}'s ability to guide a user through a hypothetical molecular discovery workflow.
As exemplified in Appendix~\autoref{fig:workflow} via the common herbicide Monuron, three consecutive calls to \emph{Text+Chem T5} are sufficient to convert a textual description of Monuron to a stepwise wet-lab execution protocol to synthesize it.
None of the three steps -  (1) \texttt{text2mol} -- converting the text description to SMILES; (2) \texttt{retro} -- converting the SMILES to a precursor list for the synthesis and (3) \texttt{paragraph2actions} -- converting a patent description of the generated reaction to a stepwise synthesis execution protocol - can be solved by SOTA LLMs like ChatGPT~\cite{ouyang2022training} or Galactica~\cite{taylor2022galactica}\footnote{Note that the patent description was retrieved manually}. 
Overall, Galactica failed to provide SMILES strings even though it was developed for scientific text; ChatGPT made small errors in all tasks and only \emph{Text+Chem T5} was able to arrive at the correct result. 
See Appendix~\ref{appx:comparison-workflow} for more comparisons and examples.
\label{section:related-work}

\section{Related Work}
\paragraph{Transformers for Natural Language.}
BERT~\citep{devlin2018bert} is a prominent example, a bidirectional transformer trained using masked language modeling, which forces the model to reconstruct the input that has been degraded. 
BERT popularized the adoption of the transformer as the backbone architecture for many language tasks.
In the same period, consistent, long-range text generation was achieved using autoregressive training (causal language modeling) using the GPT-family~\citep{radford2018improving}, transformers with a specific focus on the next token generation.
Surprisingly, simple training using large architectures and vast data gave rise to generalization, multitasking~\citep{radford2019language}, and few-shot capabilities~\citep{brown2020language}.
Training on code~\citep{chen2021evaluating}, instruction finetuning~\citep{ouyang2022training, chung2022scaling} and chain-of-thought prompting~\citep{wei2022chain} have further improved the performance of transformer models in reasoning tasks~\citep{fu2022complexity, lewkowycz2022solving, lievin2022can}.

\textbf{Generative Transfer and Multi-task Learning.}
Building on the success of masked language modelling~\citep{devlin2018bert}, the T5 framework~\citep{raffel2020exploring} has emerged as a leading paradigm in generative transfer and multitask learning. T5 is a multitask language model that is trained on a wide variety of tasks using masking. This model has shown impressive generalization and adaptation capabilities for multimodal generation~\citep{saharia2022photorealistic}. 
These techniques have also greatly improved the performance of cross-domain language models, such as models that can translate between languages~\citep{sanh2021multitask}.

\textbf{Language Models in Chemistry.}
Recent advances in neural language models have been successfully applied to the chemical domain. Several studies have utilized transformer architectures to address a variety of chemical tasks, including forward reaction prediction~\citep{schwaller2019molecular}, multi-step retrosynthesis~\citep{schwaller2020predicting, toniato2021unassisted}, and property prediction~\citep{schwaller2021mapping, vaucher2020automated, born2022regression}.
In addition, there has been research exploring multitask generation for chemistry using a pretrained T5 model with multiple heads for different types of tasks, such as regression, classification, and generation~\citep{lu2022unified}.
However, there is still a need for models that can handle multiple tasks across domains without the need for expensive pretraining or finetuning.
\label{section:conclusion}

\section{Limitations}
This study introduces a novel application of large language models in chemistry and cross-domain generation, which is subject to the same limitations as previous research on the topic~\cite {edwards2022translation} and the presence of unintended biases encoded in the training data~\cite{raffel2020exploring, liang2021towards}. The chemistry representation is based on SMILES~\cite{weininger1988smiles}, which can generate invalid sequences. Alternative representations like SELFIES with validity guarantees could improve this issue~\cite{krenn2020self}. 
Finally, it is worth noting that all the presented models were designed for research purposes only, and any molecules generated by the model should undergo standard clinical testing before being used for medical or other purposes.

\section{Conclusion}
In this paper, we introduced \textit{Multitask Text and Chemistry T5}, a multi-task, multi-domain language model for the natural and chemical domains. The model can effectively translate between natural and chemical languages, making it possible to solve a variety of tasks such as chemical reaction prediction, retro-synthesis, text-conditional de novo generation, and molecular captioning. The strength of the model lies in its ability to solve multiple tasks without the need for additional task-specific heads or adaptation at test time. The result is a step forward in the development of a general multi-task, multi-domain language model for the life sciences. This can potentially accelerate the discovery process in this field, by providing a more efficient way to process, analyze and generate chemical and textual data.

\section*{Code Availability}

The \emph{Multitask Text and Chemistry T5} model is available for inference, training and finetuning via the GT4SD library~\cite{manica2022gt4sd}:~\url{https://github.com/GT4SD/gt4sd-core}.

A gradio~\cite{abid2019gradio} app, build on top of the GT4SD implementation and hosted publicly on HuggingFace spaces, allows easy access to the models:~\url{https://huggingface.co/spaces/GT4SD/multitask-text-and-chemistry-t5}.

Code is available at:~\url{https://github.com/GT4SD/multitask_text_and_chemistry_t5}

\small
\bibliographystyle{apalike}
\bibliography{biblio.bib}

\newpage
\normalsize
\appendix
\onecolumn
\section{Additional Experiments}

\begin{table}[ht]
    \centering
    \caption{\textbf{Comparison on the cross-domain tasks}. 
    The models include a fine-tuned transformer, T5 models in both zero-shot and fine-tuned settings, MolT5, and Text+Chem T5. The models are compared at two different sizes, "small" and "base." We can see how T5 zero-shot (without fine-tuning) is completely unable to perform cross-domain translation, corroborating the necessity for multi-domain multitask modelling in the chemical and natural language domains.
    }
    \begin{tabular}{c c | cc}
    \toprule
    & Size & \texttt{text2mol} & \texttt{mol2text} \\
    \midrule
    Transformer (fine-tuned) & - & 0.499 & 0.061  \\
    \underline{T5 (zero-shot)}    & small & 0.000  & 0.004 \\
    T5 (fine-tuned)    & small & 0.762  & 0.501 \\
    MolT5             & small & 0.755 & 0.519  \\
    \textbf{\emph{Text+Chem T5}} & small & \textbf{0.815} & \textbf{0.560} \\
    \midrule
    \underline{T5 (zero-shot)}    & base & 0.000  & 0.003 \\
    T5 (fine-tuned)    & base & 0.762  & 0.511 \\
    MolT5             & base & 0.769 & 0.540  \\
    \textbf{\emph{Text+Chem T5}} & base & \textbf{0.853} & \textbf{0.625} \\
    \bottomrule
    \end{tabular}
    \label{tab:experiments-appendix}
\end{table}

\begin{table}[ht]
    \centering
    \caption{\textbf{Results of the Paragraph to Actions task}. 
    The performance of the models is evaluated by BLEU score and accuracy. The results show that the proposed model (Text+Chem T5) outperforms other baselines in all metrics. These results demonstrate the effectiveness of the proposed model in the text modality. RXN model is the paragraph-to-action proposed in~\citep{vaucher2020automated}.
    }
    \begin{tabular}{cc | cc}
    \toprule
     & Size & BLEU score $\uparrow$ &Accuracy $\uparrow$ \\
    \midrule
    RXN & - & 0.850 & 0.608\\
    T5 (fine-tuned) & small & \textbf{0.953} & \textbf{0.856} \\
    \textbf{\emph{Text+Chem T5}} & small &0.929&0.780 \\
    \textbf{\emph{Text+Chem T5-augm}} & small & 0.926&0.780 \\
    \midrule
    \textbf{\emph{Text+Chem T5}} & base & 0.935 & 0.800  \\
    \textbf{\emph{Text+Chem T5-augm}} & base & 0.943 & 0.829  \\
    \bottomrule
    \end{tabular}
    \label{tab:experiments-paragraph-to-actions}
\end{table}

\begin{figure}[ht]
    \centering
\includegraphics[width=.75\columnwidth, keepaspectratio]{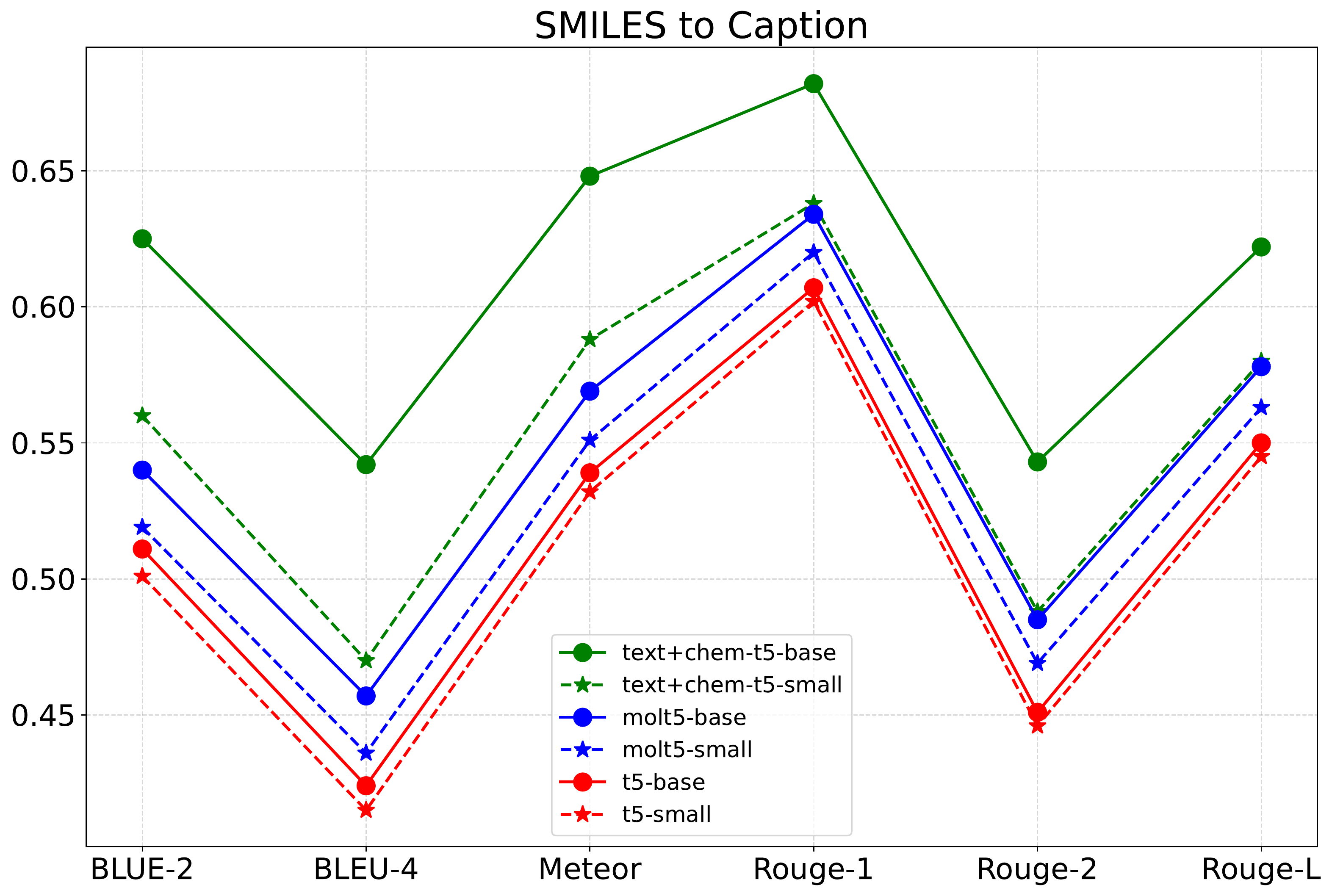}
    \caption{\textbf{Molecule to Caption task}.
    This plot compares the performance of three different models with different sizes (Text+Chem T5-base, Text+Chem T5-small, MolT5-base, MolT5-small, T5-base, and T5-small) on the task of converting SMILES to captions, using six different metrics: BLUE-2, BLEU-4, Rouge-1, Rouge-2, Rouge-L, and Meteor. The models are compared by plotting their scores on the y-axis. The graph shows that our proposal, Text+Chem T5, performs the best on all metrics and improves with size, corroborating our hypothesis that joint learning on molecular and textual domains leveraging multitask learning is a powerful paradigm to bridge the gap between domains.}
    \label{fig:mol2text_figure1-appx}
\end{figure}

\begin{figure}[ht]
    \centering
\includegraphics[width=.75\columnwidth, keepaspectratio]{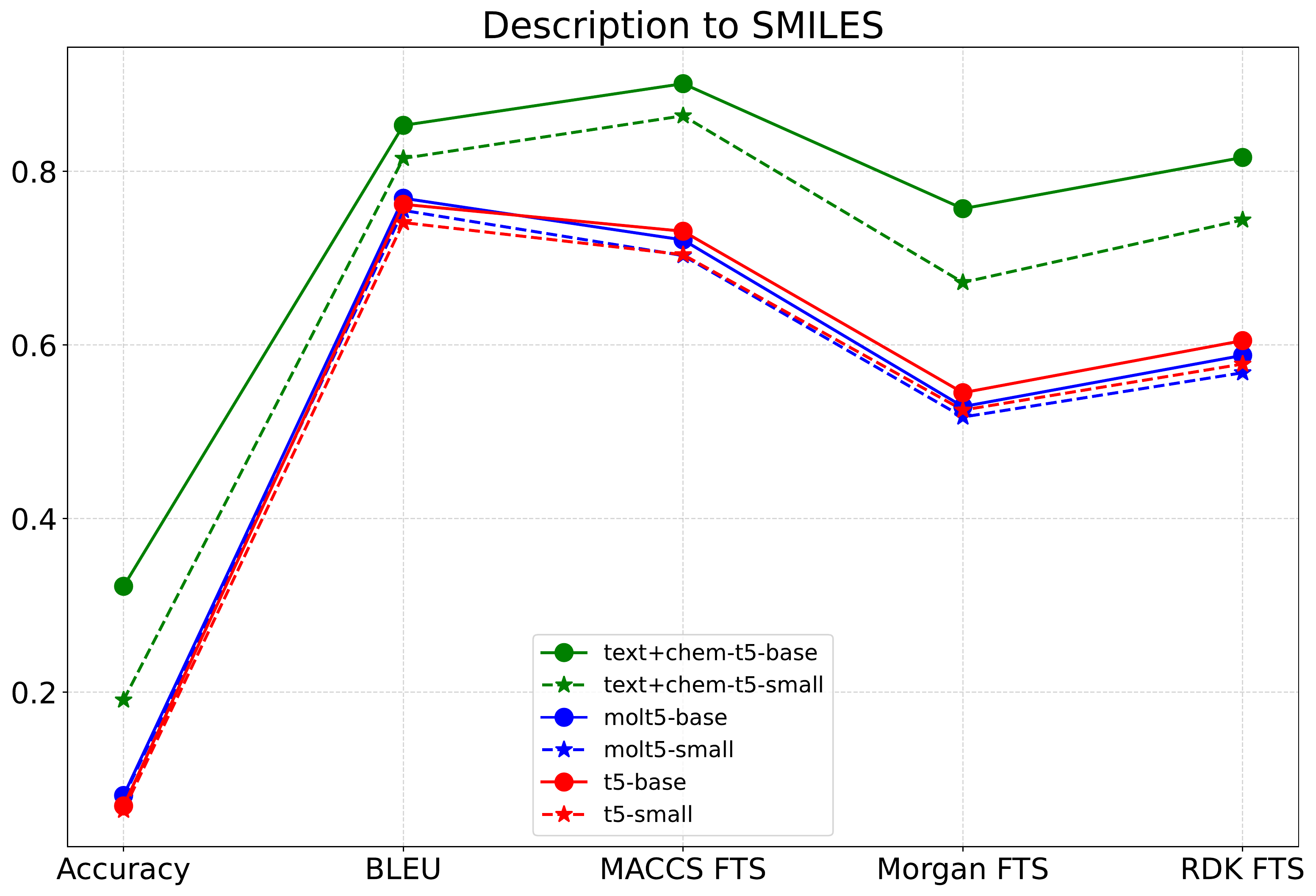}
    \caption{\textbf{Text-conditional Molecule Generation}.
    This plot compares the performance of three different models with different sizes (Text+Chem T5-base, Text+Chem T5-small, MolT5-base, MolT5-small, T5-base, and T5-small) on the task of converting captions to SMILES, using five different metrics: Accuracy, Morgan FTS, RDK FTS, BLEU, MACCS FTS. The models are compared by plotting their scores on the y-axis. The graph shows that our proposal, Text+Chem T5, performs the best on all metrics and improves with size, corroborating our hypothesis that joint learning on molecular and textual domains leveraging multitask learning is a powerful paradigm to bridge the gap between domains.}
    \label{fig:text2mol_figure1-appx}
\end{figure}

\clearpage
\section{Comparison with recent Language Models and Workflow example}
\label{appx:comparison-workflow}
\begin{table}[h!]
\centering
\small
\caption{\textbf{Conditional Molecule Generation}. 
Given a description, generate the SMILES representation. We compare the generation of our model with GALACTICA (1.3B) and ChatGPT.
For our model and ChatGPT we use the prompt structure proposed in Table~\ref{app:prompts}. For GALACTICA, we follow the prompt structure proposed by the authors.}
\resizebox{0.99\textwidth}{!}{
\begin{tabular}{c | p{12cm} }
&  \\
\toprule
Input  & The molecule is an acyl-CoA oxoanion
that is the pentaanion of (S)-3-hydroxydecanedioyl-CoA arising from deprotonation of the phosphate, diphosphate and carboxylic acid functions; major species at pH 7.3. It is a conjugate base of a (S)-3-hydroxydecanedioyl-CoA. \\\\
Target & CC(C)(COP(=O)([O-])OP(=O)([O-])OC[C@@H]1[C@H]([C@H]([C@@H](O1) N2C=NC3=C(N=CN=C32)N)O)OP(=O)([O-])[O-])[C@H](C(=O)NCCC(=O) NCCSC(=O)C[C@H](CCCCCCC(=O)[O-])O)O \\\\
\hdashline \\
GALACTICA & (S)-3-hydroxydecanedioyl-CoA \\\\
ChatGPT & [O-]P(=O)(O[C@@H]1CCC@@([NH3+])C1)OCC(=O)NCCCC@HC \\\\
Text+Chem T5 (Ours) & 
CC(C)(COP(=O)([O-])OP(=O)([O-])OC[C@@H]1[C@H]([C@H]([C@@H](O1) N2C=NC3=C(N=CN=C32)N)O)OP(=O)([O-])[O-])[C@H](C(=O)NCCC(=O) NCCSC(=O)C[C@H](CCCCCCC(=O)[O-])O)O \\\\
&  \\
\toprule
Input  & The molecule is a glucotriose consisting of D-glucopyranose, alpha-D-glucopyranose and beta-D-glucopyranose residues joined in sequence by two (1->4) glycosidic bonds. The configuration of the anomeric centre at the non-reducing terminus is not specified. It is a partially-defined glycan and a glucotriose. \\\\
Target &  C([C@@H]1[C@H]([C@@H]([C@H](C(O1)O[C@@H]2[C@H](O[C@@H]([C@@H] ([C@H]2O)O)O[C@@H]3[C@H](O[C@H]([C@@H]([C@H]3O)O)O)CO)CO)O)O)O)O \\\\
\hdashline \\
GALACTICA & (a) (b) (c) (d) (e) (f) (g) (h) \\\\
ChatGPT & O[C@H]1C@@HC@HC@@HC@@H[C@H]1O \\\\
Text+Chem T5 (Ours) & C([C@@H]1[C@H]([C@@H]([C@H]([C@@H](O1)O[C@@H]2[C@H](O[C@@H]([C@@H] ([C@H]2O)O)O[C@@H]3[C@H](O[C@H]([C@@H]([C@H]3O)O)O)CO)CO)O)O)O)O 
 \\\\
&  \\
\toprule
Input  & The molecule is a secondary amino compound that is 3,4-dimethoxyphenylethylamine in which one of the hydrogens attached to the nitrogen has been replaced by a 4-cyano-4-(3,4-dimethoxyphenyl)-5-methylhexyl group. It is an aromatic ether, a nitrile, a polyether and a secondary amino compound. \\\\
Target &  CC(C)C(CCCNCCC1=CC(=C(C=C1)OC)OC)(C\#N)C2=CC(=C(C=C2)OC)OC \\\\
\hdashline \\
GALACTICA & 3,4-Dimethoxyphenylethylamine \\\\
ChatGPT & CN(CCCCC(C\#N)c1ccc(OC)cc1OC)CCc1ccc(OC)cc1OC \\\\
Text+Chem T5 (Ours) & CC(C)C(CCCNCC(C1=CC(=C(C=C1)OC)OC)OC)(C\#N)C2=CC(=C(C=C2)OC)OC 
\label{app:example-d2m}
\end{tabular}
}
\end{table}

\begin{table}[h!]
\centering
\small
\caption{\textbf{Molecular Captioning}. 
Given a SMILES representation generate a description in natural language. We compare the generation of our model with GALACTICA (1.3B) and ChatGPT.
For our model and ChatGPT we use the prompt structure proposed in Table~\ref{app:prompts}. For GALACTICA, we follow the prompt structure proposed by the authors.}
\resizebox{0.99\textwidth}{!}{
\begin{tabular}{c | p{12cm} }
&  \\
\toprule
Input  & C[C@]12CCC(=O)C=C1CC[C@@H]3[C@@H]2C(=O)C[C@]$\backslash$$\backslash$4([C@H]3CC/C4=C/C(=O)OC)C \\\\
Target &  
The molecule is a steroid ester that is methyl (17E)-pregna-4,17-dien-21-oate substituted by oxo groups at positions 3 and 11. It is a 3-oxo-Delta(4) steroid, an 11-oxo steroid, a steroid ester and a methyl ester. It derives from a hydride of a pregnane. \\\\
\hdashline\\
GALACTICA &  -C(=O)OC \\\\
ChatGPT &  The SMILES representation corresponds to the chemical structure of a steroid molecule called cortisone. A possible caption could be: "SMILES representation of cortisone, a steroid hormone involved in various physiological processes in the body."\\\\
Text+Chem T5 (Ours) & The molecule is an androstanoid that is the methyl ester of cortisone. It has a role as a metabolite. It is a cortisol ester, an androstanoid, a methyl ester and a 3-oxo-Delta(4) steroid. It derives from a cortisone. 
 \\\\
&  \\
\toprule
Input  &  C(C(C(=O)O)N)S\\\\
Target &  The molecule is a sulfur-containing amino acid that is propanoic acid with an amino group at position 2 and a sulfanyl group at position 3. It has a role as a fundamental metabolite. It is an alpha-amino acid, a polar amino acid and a sulfur-containing amino acid. It contains a sulfanylmethyl group. It is a conjugate base of a cysteinium. It is a conjugate acid of a cysteinate(1-). It is a tautomer of a cysteine zwitterion.       
\\\\
\hdashline\\
GALACTICA &  C  \\\\
ChatGPT & The SMILES representation corresponds to the chemical structure of a molecule called cysteine. A possible caption could be: "SMILES representation of cysteine, an amino acid that is a building block of proteins and plays important roles in various biological processes." 
\\\\
Text+Chem T5 (Ours) & 
The molecule is a cysteinyl radical derived from cystein. It has a role as a fundamental metabolite. It is a cysteinyl radical and a non-proteinogenic alpha-amino acid. It derives from a cystein. It is a conjugate base of a cysteinyl radical cation. It is a conjugate acid of a cysteinyl radical. It is a tautomer of a cysteinyl radical zwitterion.
 \\\\
 &  \\
\toprule
Input  & CC(=O)NC(CC1=CC=C(C=C1)O)C(=O)O \\\\
Target &  
The molecule is an N-acetyl-amino acid that is tyrosine with an amine hydrogen substituted by an acetyl group. It has a role as a human urinary metabolite. It is a tyrosine derivative, a N-acetyl-amino acid and a member of phenols. It derives from a tyrosine.
\\\\
\hdashline\\
GALACTICA & CC(=O)NC(CC1=CC=C(C=C1)O)C(=O)O \\\\
ChatGPT  &
The SMILES representation corresponds to the chemical structure of a molecule called acetaminophen, which is a common pain reliever and fever reducer. A possible caption could be: "SMILES representation of acetaminophen, a widely used analgesic and antipyretic medication."
\\\\
Text+Chem T5 (Ours) & 
The molecule is the N-acetyl derivative of N-acetyltyrosine. It derives from a N-acetyltyrosine. It is a conjugate acid of a N-acetyltyrosinate.
\label{app:example-m2d}
\end{tabular}
}
\end{table}

\begin{figure}[ht]
    \centering
\includegraphics[width=\columnwidth, keepaspectratio]{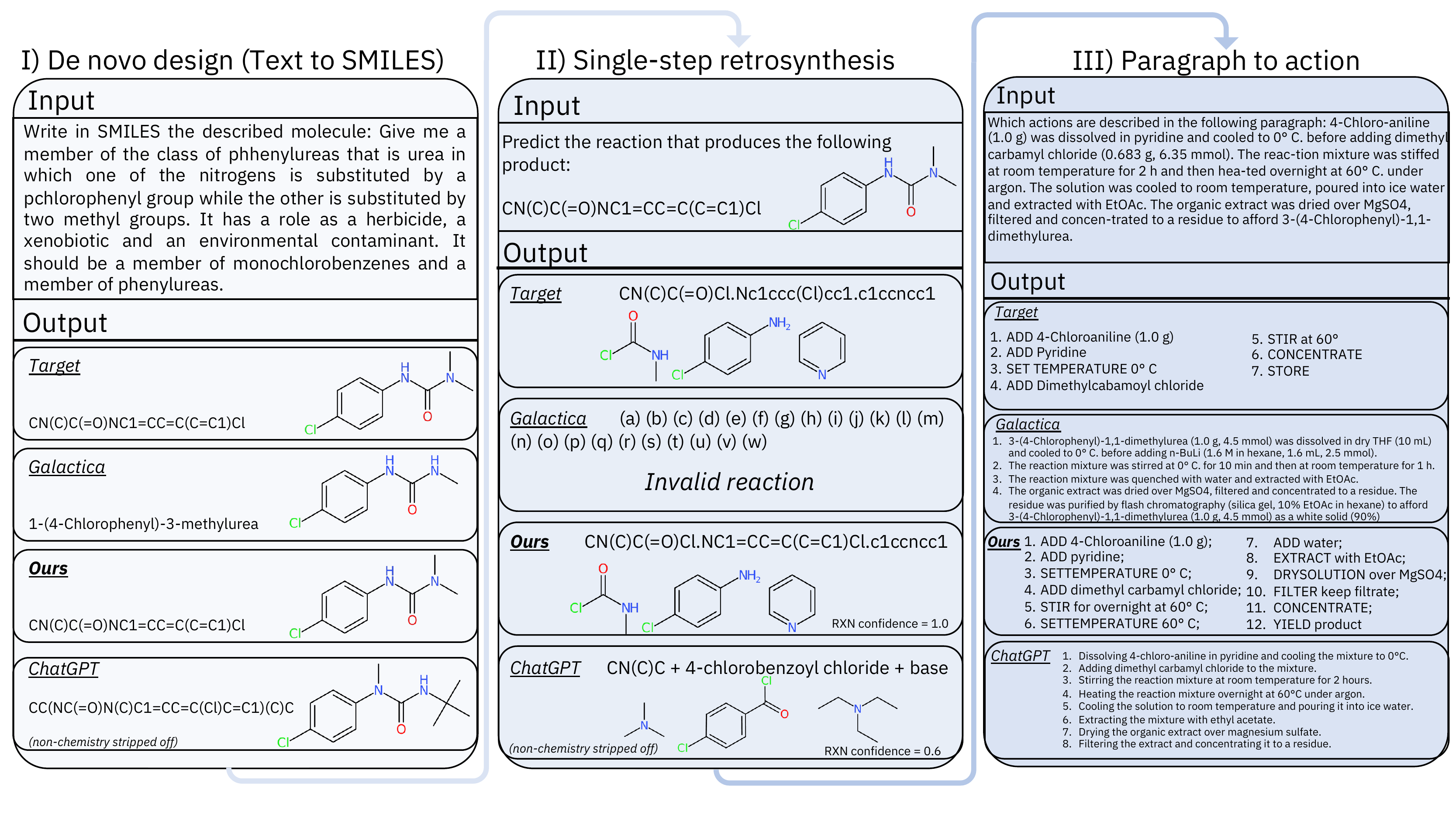}
\caption{\textbf{Discovery workflow}. 
Qualitative comparison of \emph{Text+Chem T5} to \textit{Galactica} 1.3B~\cite{taylor2022galactica} and \textit{ChatGPT} for a hypothetical discovery workflow of Monuron, a commonly used herbicide. The discovery workflow starts from a textual description of Monuron and goes all the way to a stepwise execution protocol for wet-lab synthesis. \newline
\textbf{I) }Text to SMILES task: Only \textit{Text+Chem T5} produces exactly the desired molecule. \textit{Galactica} fails to provide a valid SMILES string (when forced to do so), but gives a textual description of a very similar molecule. \textit{ChatGPT} also produces a similar molecule but adds to many methyl groups to the urea nitrogen. \textit{ChatGPT} also generated a verbose description of each block of the SMILES string. \newline
\textbf{II) }Retrosynthesis: Asking the model to find a synthetic route for Monuron. While \textit{Galactica} fails to give a sensible output, \textit{ChatGPT} hesitantly suggested a possible route: "One possible reaction that could lead to the formation of this product is the reaction between an amine (e.g. N,N-dimethylamine) and a substituted benzoyl chloride (e.g. 4-chlorobenzoyl chloride) in the presence of a base (e.g. triethylamine), which would result in the formation of an amide with the same substituents as the product". 
This reaction is chemically valid, but unlikely to succeed, as the forward reaction prediction models from IBM RXN for Chemistry~\cite{schwaller2019molecular} only assign a confidence of $0.6$. On the contrary, the reaction generated by \textit{Text+Chem T5} is identical to the target reaction and is predicted to succeed by RXN with a confidence of $1.0$.
\newline
\textbf{III)} Paragraph to actions: Last, one might be interested to execute the proposed reaction in a chemistry lab. Since the identified reaction is not heavily documented in the literature, we utilized a reaction atlas~\cite{schwaller2021mapping} to identify an extremely similar reaction (\url{https://patents.google.com/patent/US8697887}; shown in~\autoref{fig:simreaction}). This reaction has a literature-reported experimental procedure that has been validated and patented. We adapted this procedure to arrive at the prompt for the third task.
All models split up the procedure in individual steps and \textit{Text+Chem T5} as well as \textit{ChatGPT} conceptually succeeded at this task. Instead, \textit{Galactica} heavily invented information, even mixing the ingredients (e.g., 3-(4-Chlorophenyl)-1,1-dimethylurea mentioned in step one as an ingredient is the IUPAC name for Monuron, our desired product of the reaction).  
\newline
Generally, we use the prompt structure proposed in Table~\ref{app:prompts} for all models, while for GALACTICA, add "ANSWER: " to the original prompt, as proposed by the authors. Note that the shown images were not generated by the model, but rendered from the SMILES string or the textual description (\textit{Galactica}).
}
    \label{fig:workflow}
\end{figure}

\begin{figure}[ht]
    \centering
\includegraphics[width=.75\columnwidth, keepaspectratio]{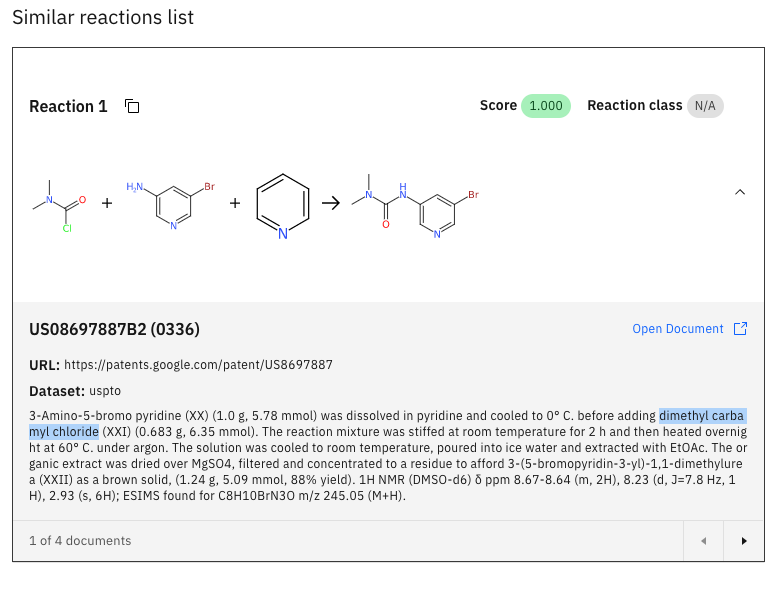}
    \caption{
    \textbf{Reaction most similar to the generated synthesis route for Monuron.}.
    Reaction identified using \textit{IBM RXN for Chemistry} (\url{https://rxn.res.ibm.com/rxn/}) using the reaction atlas described in~\citet{schwaller2021mapping}. The reaction is identical to the identified synthesis route for Monuron apart from the 3-Amino-5-bromo pyridine which is substituted with 4-Chloroanyline to yield Monuron.
    }
\label{fig:simreaction}
\end{figure}

\clearpage

\section{Prompt templates}

\begin{table}[h!]
\centering
\caption{\textbf{Prompt templates} that have been used for the 5 tasks of interest for our multi-task and multi-domain training.}
\begin{tabular}{c|c}
Task                  & Template            \\ 
\toprule
Forward prediction    & Predict the product of the following reaction: \textless{}input\textgreater{}            \\
Retrosynthesis        & Predict the reaction that produces the following product: \textless{}input\textgreater{} \\
Paragraph-to-actions  & Which actions are described in the following paragraph: \textless{}input\textgreater{}   \\
Description-to-smiles & Write in SMILES the described molecule: \textless{}input\textgreater{}                   \\
Smiles-to-caption     & Caption the following SMILES: \textless{}input\textgreater{}
\label{app:prompts}
\end{tabular}
\end{table}

\section{Experimental Details}
\label{app:details}

\begin{table}[ht]
    \centering
    \caption{Models Size.}
    \begin{tabular}{c c | c c c}
    \toprule
         Model & Suffix  & Parameters (M) & Multi-task & Multi-domain \\
         \midrule
         RXN &  forward           & 12  & \xmark & \xmark \\
         RXN &  retrosynthesis    & 12  & \xmark & \xmark \\
         RXN &  paragraph2action  & 20  & \xmark & \xmark \\
         \midrule
         T5           & small     & 60  & \cmark & \xmark \\
         Text+Chem T5 & small     & 60  & \cmark & \cmark \\
         Transformer  &   -       &  65 & \cmark & \xmark \\
         MolT5        & small     & 77  & \xmark & \cmark \\
         \midrule
         T5           & base      & 220 & \cmark & \xmark \\
         Text+Chem T5 & base      & 220 & \cmark & \cmark \\
         MolT5        & base      & 250 & \xmark & \cmark \\
         \bottomrule
    \end{tabular}
    \label{tab:model-size}
\end{table}

\begin{table}[ht]
\begin{center}
\caption{Relevant Hyperparameters for Text+Chem T5.}
\resizebox{0.99\textwidth}{!}{
\begin{tabular}{lcccc} 
\toprule
& Text+Chem T5-small & Text+Chem T5-base & Text+Chem T5-small augm & Text+Chem T5-base augm \\ 
\midrule
Dataset                 & standard & standard & augmented & augmented \\
Pretrained base         & t5-small   & t5-base        & t5-small & t5-base \\
Heads                   &  8        &   12           & 8   & 12         \\
Layers                  &  6         &   12            & 6    & 12          \\
Parameters              & 60M & 220M & 60M & 220M \\
Input max length        & 512        & 512            & 512      & 512 \\
Epochs                  & 1 & 1 & 1 & 1 \\
Batch size              &  64        &  64            & 64       & 64     \\
Accumulated gradient batches & 4 & 4 & 4 & 4 \\
Learning rate           &  $4e^{-4}$        &  $6e^{-4}$        & $6e^{-4}$ & $6e^{-4}$ \\
\bottomrule
\end{tabular}
}
\end{center}
\label{tab:params}
\end{table}

\end{document}